\documentclass[10pt,twocolumn,letterpaper]{article}

\usepackage[pagenumbers]{cvpr} % To force page numbers, e.g. for an arXiv version

\usepackage[dvipsnames]{xcolor}

\definecolor{cvprblue}{rgb}{0.21,0.49,0.74}
\usepackage[pagebackref,breaklinks,colorlinks,citecolor=cvprblue]{hyperref}

\def\paperID{*****} % *** Enter the Paper ID here
\def\confName{}
\def\confYear{2024}

\title{Fish Tracking Challenge 2024: \\
A Multi-Object Tracking Competition with Sweetfish Schooling Data}

\author{Makoto M. Itoh\\
Shitennoji University
\and
Qingrui Hu\\
Nagoya University
\and
Takayuki Niizato\\
University of Tsukuba
\and
Hiroaki Kawashima\\
University of Hyogo 
\and
Keisuke Fujii\\
Nagoya University
}
\begin{document}
\maketitle

\begin{abstract}
The study of collective animal behavior, especially in aquatic environments, presents unique challenges and opportunities for understanding movement and interaction patterns in the field of ethology, ecology, and bio-navigation. The Fish Tracking Challenge 2024 (\url{https://ftc-2024.github.io/}) introduces a multi-object tracking competition focused on the intricate behaviors of schooling sweetfish. Using the SweetFish dataset, participants are tasked with developing advanced tracking models to accurately monitor the locations of 10 sweetfishes simultaneously. This paper introduces the competition's background, objectives, the SweetFish dataset, and the appraoches of the 1st to 3rd winners and our baseline. By leveraging video data and bounding box annotations, the competition aims to foster innovation in automatic detection and tracking algorithms, addressing the complexities of aquatic animal movements. The challenge provides the importance of multi-object tracking for discovering the dynamics of collective animal behavior, with the potential to significantly advance scientific understanding in the above fields.
\end{abstract}

\section{Introduction}
\label{sec:intro}
Collective animal behaviors are teeming with life and intricate behavioral patterns. Fish schooling behavior offers a unique window into understanding animal navigation in water. For ethologists, ecologists, and mathematical and theoretical biologists, decoding these patterns is important. However, automatically tracking the movement of fishes, especially when in schools, introduces many challenges. 

By developing advanced tracking platform \cite{walter2021trex,pereira2020sleap,mathis2018deeplabcut} and models (e.g., \cite{wojke2017simple,zhang2022bytetrack}), researchers can uncover the intricacies of aquatic movement and significantly advance this field. Originally, observation relied on the human eye \cite{tinbergen1963}, but recent technological innovations have fostered the increase of observational methodologies employing digital tools \cite{fernandes2020}. Utilizing digital video cameras facilitates objective and comprehensive observation, surpassing human visual capabilities, enabling simultaneous observation over wide areas \cite{bonelli2020}. Furthermore, the utilization of various recording devices such as night vision cameras, thermography cameras, sonar cameras, super slow-motion cameras, and drone cameras enables the observation of phenomena imperceptible to human visual inspection \cite{dell2014}. It is anticipated that observation methodologies leveraging digital equipment will continue to expand in the future.

The primary aim of this study is to advance the understanding and analysis of collective animal behavior in aquatic environments through the development and application of innovative multi-object tracking (MOT) models. By focusing on the intricate behaviors of schooling sweetfish, the study seeks to address and overcome the challenges associated with accurately monitoring and analyzing the movement and interaction patterns of aquatic animals in groups. The Fish Tracking Challenge 2024 (\url{https://ftc-2024.github.io/}), utilizing the comprehensive SweetFish dataset, provides a unique platform for researchers and technologists to develop, test, and refine advanced tracking algorithms capable of high-fidelity monitoring of multiple fish simultaneously.

This endeavor is not only important for the fields of ethology, ecology, and bio-navigation but also sets a precedent for interdisciplinary collaboration in the pursuit of understanding complex biological systems. The competition's emphasis on leveraging video data and bounding box annotations to foster innovation in automatic detection and tracking algorithms aims to catalyze breakthroughs in how we approach the study of collective animal behavior. Ultimately, the study's purpose is to enhance our scientific understanding of aquatic animal movements, contributing to broader applications in environmental conservation, sustainable fisheries management, and the development of autonomous navigation systems inspired by biological systems.

The remainder of this paper is organized as follows. First, in Section \ref{sec:dataset}, we describe our Sweetfish dataset used in the competition. Next, we describe our baseline and competition winners' methods in Section \ref{sec:methods}.
We then present competition results in Section \ref{sec:results}, and conclude this paper in Section \ref{sec:conclusion}.

\section{Dataset and evaluation}
\label{sec:dataset}
In this competition, the dataset in the previous work \cite{niizato2024information} was used.
The ayu or sweetfish ({\it{Plecoglossus altivelis}}) was collected, which are widely farmed throughout Japan. Juvenile ayus (approximately 7–14 cm in body length) shows typical schooling behavior. The experimental arena comprised a 3 $\times$ 3 $m^2$ shallow white tank. The water depth was approximately 15 cm (i.e., the schools were approximately two-dimensional). In the competition dataset, the spatial resolution of video was 2456 $\times$ 2048 pixels and a temporal resolution was 15 frames per second.  
The study \cite{niizato2024information} was conducted in strict accordance with the recommendations of the Guide for the Care and Use of Laboratory Animals from the National Institute of Health. The study protocol was approved by the Committee on the Ethics of Animal Experiments at University of Tsukuba (Permit Number: 14-386). 
All efforts were made to minimize animal suffering. 

For each frame, the center point of each sweetfish was annotated. To adapt the dataset for the MOT task, the average bounding box size was calculated and applied to all annotations.
The dataset was split into training, development, and test sets to facilitate model evaluation and generalization. The training set is used to train the model (i.e., bounding boxes are given), the development set is used to fine-tune the hyperparameters and to confirm the submission results (i.e., bounding boxes are given), and the test set is used to evaluate the final model performance (bounding boxes are not given).
In total, the sweetfish dataset consists of 165,150 annotated bounding boxes of 10 sweetfishes. The dataset is divided into training (5 min 33 sec), development (1 min 15 sec), and test (11 min 33 sec) sets, with a total duration of 18 min 21 sec.
 
\begin{figure}[h]
\centering
\includegraphics[width=0.5\textwidth]{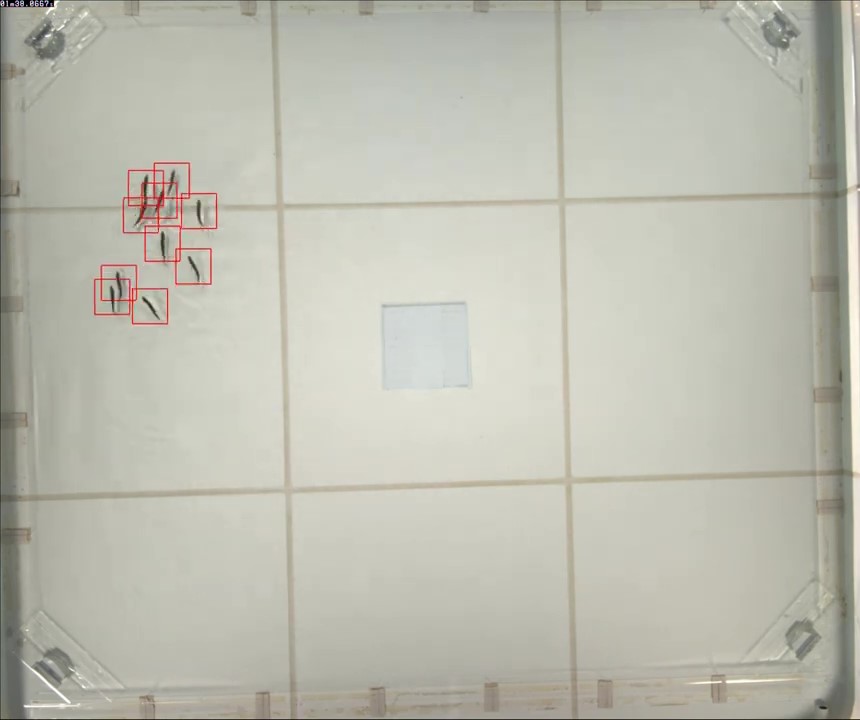}
\caption[]{\textcolor{black}{Sweetfishes and their bounding boxes in the Sweetfish dataset.}
}
\label{fig:sweetfish}
\end{figure}
 
The goal of this challenge is accurate tracking of 10 sweetfishes. Performance of the models are be evaluated based on HOTA (Higher Order Tracking Accuracy) score, which is a holistic and popular score in MOT. HOTA is designed to overcome many of the limitations of previous metrics \cite{luiten2021hota}. 
% Hu
HOTA consists of detection accurary (DetA), localization accuracy (LocA), and association accuracy (AssA), the metrics combine the evaluation of detection accuracy, tracking accuracy, and false positives (FPs). HOTA finds a better balance between these two extremes by equally weighting both detection and association, while allowing analysis of each component separately with the detection accuracy and association accuracy sub-scores. In the competition results, the number of ID swithes (IDs) and false negatives (FNs) are also indicated for a reference.
 
As a condition to being awarded prizes, top-3 winners fulfilled the following obligations. After the final submission deadline, they submited their code so that the organizers can check for cheating.
And they submited short report papers that describe the awarded methodologies \cite{tachioka2024,wang2024,gao2024}.

\section{Methods}
\label{sec:methods}
In this section, we describe the methodologies by us (baseline method) and by the top participants of the Fish Tracking Challenge 2024, each utilizing the YOLOv8 object detector \cite{Jocher_YOLO_by_Ultralytics_2023} and various tracking algorithms to address the complex task of sweetfish tracking.
Here, we describe the methodologies from the baseline approach to the strategies employed by the first through third-place winners.

% Hu
\subsection{Baseline appraoch}
As a baseline, we chose YOLOv8 \cite{Jocher_YOLO_by_Ultralytics_2023} as our object detector to achieve real-time and high-accuracy detection performance. We used the officially provided yolov8l pre-training model to train our model with the SweetFish dataset, and the first 7000 frames of the training video were used as the training set, the next 3000 frames were the test set. We trained the model over 200 epochs. We used the SGD optimizer with a weight decay of 0.0005 and momentum of 0.937 (default parameters). The initial learning rate was 0.01, and the batch size was set to 32. All the experiments were conducted on a single Nvidia RTX 4090 GPU.

After training the detector, we used the ByteTrack \cite{zhang2022bytetrack} to track sweetfish. 
Since most fish have a confidence level of about 0.9, we set the detection threshold a little loosely. Thus the threshold for detection to be treated as high-score detection was 0.5, while detections with a confidence score between 0.5 and 0.1 was treated as low-score detections, and the rest of detections with a confidence score lower than 0.1 was filtered. The detection threshold of a new track was set to 0.6, and the unmatched detection was 0.8. The max frame for keeping lost tracks was 30. 
% Hu

\subsection{Approach of the 1st place \cite{tachioka2024}}
The participant trained a YOLOv8 detector \cite{Jocher_YOLO_by_Ultralytics_2023}  for fish detection starting from the pre-trained model yolov8x. The number of training epochs was 30 and if the model was not updated in 10 epochs, training stopped. Other settings are described in \cite{tachioka2024}.

After detection, Deep OC-SORT \cite{DeepOCSORT} was used as a tracker, although the ReID model was ineffective for fish detection due to the similarity of individuals' appearance.
%because individual fish cannot be distinguished.
Deep OC-SORT has many hyperparameters which need to be adjusted to choose the best one. Properly tuning the hyperparameters is an important aspect of optimizing the model's predictive performance. In this method, fourteen parameters (see \cite{tachioka2024}) of the Deep OC-SORT tracker were tuned by using evolutionary computation performed by using the Optuna framework \cite{Optunaframework}. The HOTA was evaluated on the validation dataset by using Tree-structured Parzen Estimator (TPE) sampler \cite{TPE} and it was used for the evolutionary computation.

\subsection{Approach of the 2nd place \cite{wang2024}}
The participants implemented the YOLOv8 \cite{Jocher_YOLO_by_Ultralytics_2023} for detection. They built the network through Ultralytic API to compare two trackers, ByteTrack \cite{zhang2022bytetrack} and BoTSORT \cite{aharon2022bot}, and to change several parameters. Despite the limited number of trials, the method using ByteTrack with the confidence threshold set at 0.695 had the highest HOTA score, hence this was used in this study. 

When using ByteTrack or BoTSORT, one problem occurred \cite{wang2024}. Due to false negatives in detection, when an individual cannot be detected for a certain period of time, a different new ID is assigned to any new individuals detected thereafter. To solve this problem, they proposed Iterative Track Connector, which includes merging and interpolation, as a post-processing.

In this approach \cite{wang2024}, a distance matrix is initially constructed between all pairs of tracks, incorporating both spatial and temporal information. They compute the Euclidean distance between the locations of the last instance of one track and the first instance of another track in cases where there is no temporal overlap and the frame gap is less than MaxFrameGap, the threshold set to avoid connecting the tracks that spatially close but temporally far apart.  For all other locations on the matrix, the distance is assumed to be infinite. For each iteration, the algorithm merges one pair of tracks with the minimal distance. After every merging operation, a new distance matrix is calculated, normalized, and the process is repeated until all values on the matrix reach infinity. After all merging phase was finished, linear interpolation was employed to fill in missing bounding boxes between fish instances with the same track IDs across gap frames.

%Hu
\subsection{Approach of the 3rd place \cite{gao2024}}
In this approach, the detector was trained using the weights of the baseline model. Then they performed the tracker hyperparameter optimization on ByteTrack \cite{zhang2022bytetrack} and SORT \cite{bewley2016simple} with a simple procedure: (1) They evaluate HOTA on default hyperparameters. (2) For each hyperparameter, which needs to be modified: (a) They vary each hyperparameter while keeping other hyperparameters fixed and evaluate HOTA. (b) They identify the value of the hyperparameter with the best HOTA. (c) In the next iteration, they substitute the value of the hyperparameter with the value that yields the best HOTA. 

%There are two problems. 
They plotted the bounding boxes from the tracker from top-view and analyzed the problematic frames.
The two approaches are proposed to solve the following two problems.
(1) Rematch lost track: For example, in cases with 9 detections and 10 tracks, one track is lost and eventually deleted by the original trackers. They modified the code to rematch the lost track with the ``closest'' detection, using the Hungarian algorithm and IoU distance matrix. This means that two tracks will be assigned to one detection (i.e., ``one-to-many''). However, this will not be a problem and the tracker will assign detections to tracks correctly in the next frames. (2) Skip track creation: For example, in cases with 11 detections and 10 tracks, a new track is created for the extra detection in the original trackers. They modified the code to skip creating an additional track if the track count already reached 10.
The results for default and optimal hyperparameters for ByteTrack and SORT are shown in \cite{gao2024}.

\section{Results}
\label{sec:results}
\begin{table*}[t]
\centering
\caption{Tracking performance of baseline and top-3 methods.} 
\label{tab:performance}
\begin{tabular}{|c|c|c|c|c|c|c|c|}
\hline
         & HOTA $\uparrow$         & IDs $\downarrow$        & LocA $\uparrow$         & DetA $\uparrow$         & AssA $\uparrow$         & FN $\downarrow$          & FP $\downarrow$          \\ \hline
Baseline & \textbf{0.52} & 54          & \textbf{0.93} & \textbf{0.91} & \textbf{0.29} & \textbf{180} & \textbf{26} \\ \hline
1st place \cite{tachioka2024} & 0.49          & \textbf{49} & 0.90          & 0.87          & 0.28          & 327          & 39          \\ \hline
2nd place \cite{wang2024}& 0.47          & 72          & 0.88          & 0.80          & 0.28          & 905          & 271         \\ \hline
3rd place \cite{gao2024} & 0.44          & 65          & 0.92          & \textbf{0.91} & 0.21          & 101          & 100         \\ \hline
\end{tabular}
\end{table*}

Tracking performances of the baseline and top-3 methods on the test dataset submitted the competition system are shown in Table \ref{tab:performance}.
In summary, the baseline method achieves the best overall tracking performance. The 1st place method \cite{tachioka2024} based on HOTA score has the lowest number of IDs including baseline and the best scores among participants. The 2nd place method \cite{wang2024} shows better AssA than the 3rd place method \cite{gao2024}. The 3rd place method has better performances in the number of IDs, DetA, FNs, and FPs than the 2nd place method.
These results suggest that more effort in adjusting detector and tracker hyperparameters, rather than correcting IDs, may result in a significant improvement in overall tracking ability.

Next, we briefly discuss the results of each method.
In the 1st place method \cite{tachioka2024}, the diversity of solutions in evolutionary computation may be limited because the data size of validation dataset was too small for the exploration by evolutionary computation.
The detection was more robust than tracking and appearance change by wave on the surface of the water degraded the deep-learning-based tracking performance.
In the 2nd place method \cite{wang2024}, they found that ByteTrack \cite{zhang2022bytetrack} had a slightly better performance than BoTSORT \cite{aharon2022bot}. Although ByteTrack is designed to handle occlusion, as the video progresses, they reported that the ID switches lead to a large amount of incorrect associations. They also considered that due to the direct usage from Ultralytics package \cite{Jocher_YOLO_by_Ultralytics_2023}, their detector failed to detect the fish while ground truth for them exist for many times, leading a high number of false negatives. They achieved the score increase by performing a post-processing technique, which includes merging and interpolation. 
Regarding the 3rd place method \cite{gao2024}, for the occlusion problem, they can solve part of the ID switch problem through the ``one-to-many'' method, which is assigning a detection bounding box to two or more trajectories. For the problem of wave in fish detection, they skip the creation of additional trajectories by modifying the SORT code to keep the number of trajectories at 10, which can reduce certain missed detection and wrong detection problems.

\section{Conclusion}
\label{sec:conclusion}
In this paper, we introduced Fish Tracking Challenge 2024, a multi-object tracking competition focused on the schooling sweetfish. This paper outlines the competition's objectives, the SweetFish dataset, and the methods of baseline and participants. 
The challenge emphasize the importance of multi-object tracking for discovering the dynamics of collective animal behavior, with the potential to significantly advance scientific understanding in the above fields.
For future perspectives, to improve accuracy and robustness in the MOT task, exploring more sophisticated deep learning architectures and incorporating domain knowledges into the tracking models are considered. Regarding the extension of the current task, 3D tracking using multiple cameras and real-world aquaculture or ecological research settings can be expected.

\section{Acknowledgments}
This work is supported by JSPS KAKENHI under Grant Numbers 21H05300 and 21H05302. We would like to thank  Dr. Naoya Yoshimura at Osaka University, Atom Scott at Nagoya University, and Alex Hoi-Hang Chan at University of Konstanz for their support in holding the competition. 
{
    % \small
    \bibliographystyle{ieeenat_fullname}
    \bibliography{main}

\begin{thebibliography}{20}
\providecommand{\natexlab}[1]{#1}
\providecommand{\url}[1]{\texttt{#1}}
\expandafter\ifx\csname urlstyle\endcsname\relax
  \providecommand{\doi}[1]{doi: #1}\else
  \providecommand{\doi}{doi: \begingroup \urlstyle{rm}\Url}\fi

\bibitem[Aharon et~al.(2022)Aharon, Orfaig, and Bobrovsky]{aharon2022bot}
Nir Aharon, Roy Orfaig, and Ben-Zion Bobrovsky.
\newblock Bot-sort: Robust associations multi-pedestrian tracking.
\newblock \emph{arXiv preprint arXiv:2206.14651}, 2022.

\bibitem[Akib et~al.(2019)Akib, Sano, Yanase, Ohta, and Koyama]{Optunaframework}
Takuya Akib, Shotaro Sano, Toshihiko Yanase, Takeru Ohta, and Masanori Koyama.
\newblock Optuna: A next-generation hyperparameter optimization framework.
\newblock \emph{In Proceedings of the 25th ACM SIGKDD international conference on knowledge discovery \& data mining}, pages 2623--2631, 2019.

\bibitem[Bergstra et~al.(2011)Bergstra, Bardenet, Bengio, and Kégl]{TPE}
James Bergstra, Rémi Bardenet, Yoshua Bengio, and Balázs Kégl.
\newblock Algorithms for hyper-parameter optimization.
\newblock \emph{Advances in neural information processing systems}, 24, 2011.

\bibitem[Bewley et~al.(2016)Bewley, Ge, Ott, Ramos, and Upcroft]{bewley2016simple}
Alex Bewley, Zongyuan Ge, Lionel Ott, Fabio Ramos, and Ben Upcroft.
\newblock Simple online and realtime tracking.
\newblock In \emph{2016 IEEE international conference on image processing (ICIP)}, pages 3464--3468. IEEE, 2016.

\bibitem[Bonelli et~al.(2020)Bonelli, Melotto, Alessio~Minici, Gianfranceschi, Gobbi, Casartelli, and Caccianiga.]{bonelli2020}
Marco Bonelli, Andrea Melotto, Elena~Eustacchio Alessio~Minici, Luca Gianfranceschi, Mauro Gobbi, Morena Casartelli, and Marco Caccianiga.
\newblock Manual sampling and video observations: An integrated approach to studying flower-visiting arthropods in high-mountain environments.
\newblock \emph{Insects}, 11\penalty0 (12):\penalty0 881, 2020.

\bibitem[Dell et~al.(2014)Dell, Bender, Branson, Couzin, de~Polavieja, Noldus, Pérez-Escudero, Perona, Straw, Wikelski, and Brose]{dell2014}
Anthony~I. Dell, John~A. Bender, Kristin Branson, Iain~D. Couzin, Gonzalo~G. de Polavieja, Lucas~P.J.J. Noldus, Alfonso Pérez-Escudero, Pietro Perona, Andrew~D. Straw, Martin Wikelski, and Ulrich Brose.
\newblock Automated image-based tracking and its application in ecology.
\newblock \emph{Trends in Ecology \& Evolution}, 29\penalty0 (7):\penalty0 417--428, 2014.

\bibitem[Fernandes et~al.(2020)Fernandes, Dórea, and de~Magalhães~Rosa]{fernandes2020}
Arthur Francisco~Araújo Fernandes, João Ricardo~Rebouças Dórea, and Guilherme~Jordão de Magalhães~Rosa.
\newblock Image analysis and computer vision applications in animal sciences: An overview.
\newblock \emph{Frontiers in veterinary science}, 7:\penalty0 551269, 2020.

\bibitem[Gao et~al.(2024)Gao, Aburto, Andres, and Li]{gao2024}
Yulun Gao, Mario Aburto, Mohali Andres, and Zhongluo Li.
\newblock Report from ``labtrack team'' for fish tracking challenge 2024.
\newblock In \emph{Fish tracking challenge 2024}, 2024.

\bibitem[Hao et~al.(2024)Hao, Jiacheng, Huang, Yang, and Hwang]{wang2024}
Wang Hao, Sun Jiacheng, Hsiang-Wei Huang, Cheng-Yen Yang, and Jenq-Neng Hwang.
\newblock Fish tracking challenge 2024 report.
\newblock In \emph{Fish tracking challenge 2024}, 2024.

\bibitem[Jocher et~al.(2023)Jocher, Chaurasia, and Qiu]{Jocher_YOLO_by_Ultralytics_2023}
Glenn Jocher, Ayush Chaurasia, and Jing Qiu.
\newblock {YOLO by Ultralytics}, 2023.

\bibitem[Luiten et~al.(2021)Luiten, Osep, Dendorfer, Torr, Geiger, Leal-Taix{\'e}, and Leibe]{luiten2021hota}
Jonathon Luiten, Aljosa Osep, Patrick Dendorfer, Philip Torr, Andreas Geiger, Laura Leal-Taix{\'e}, and Bastian Leibe.
\newblock Hota: A higher order metric for evaluating multi-object tracking.
\newblock \emph{International Journal of Computer Vision}, 129:\penalty0 548--578, 2021.

\bibitem[Maggiolino et~al.(2023)Maggiolino, Ahmad, Cao, and Kitani]{DeepOCSORT}
Gerard Maggiolino, Adnan Ahmad, Jinkun Cao, and Kris Kitani.
\newblock Deep oc-sort: Multi-pedestrian tracking by adaptive re-identification.
\newblock \emph{arXiv preprint arXiv:2302.11813}, 2023.

\bibitem[Mathis et~al.(2018)Mathis, Mamidanna, Cury, Abe, Murthy, Mathis, and Bethge]{mathis2018deeplabcut}
Alexander Mathis, Pranav Mamidanna, Kevin~M Cury, Taiga Abe, Venkatesh~N Murthy, Mackenzie~Weygandt Mathis, and Matthias Bethge.
\newblock Deeplabcut: markerless pose estimation of user-defined body parts with deep learning.
\newblock \emph{Nature neuroscience}, 21\penalty0 (9):\penalty0 1281--1289, 2018.

\bibitem[Niizato et~al.(2024)Niizato, Sakamoto, ichi Mototake, Murakami, and Tomaru]{niizato2024information}
Takayuki Niizato, Kotaro Sakamoto, Yoh ichi Mototake, Hisashi Murakami, and Takenori Tomaru.
\newblock Information structure of heterogeneous criticality in fish school.
\newblock \emph{bioRxiv}, 2024.

\bibitem[Pereira et~al.(2020)Pereira, Tabris, Li, Ravindranath, Papadoyannis, Wang, Turner, McKenzie-Smith, Kocher, Falkner, et~al.]{pereira2020sleap}
Talmo~D Pereira, Nathaniel Tabris, Junyu Li, Shruthi Ravindranath, Eleni~S Papadoyannis, Z~Yan Wang, David~M Turner, Grace McKenzie-Smith, Sarah~D Kocher, Annegret~L Falkner, et~al.
\newblock Sleap: Multi-animal pose tracking.
\newblock \emph{BioRxiv}, pages 2020--08, 2020.

\bibitem[Tachioka(2024)]{tachioka2024}
Yuuki Tachioka.
\newblock Multi-object tracking with evolutionary computation based optimization for fish tracking challenge 2024.
\newblock In \emph{Fish tracking challenge 2024}, 2024.

\bibitem[Tinbergen(1963)]{tinbergen1963}
Nikolaas Tinbergen.
\newblock On aims and methods of ethology.
\newblock \emph{Zeitschrift für Tierpsychologie}, 20:\penalty0 410--433, 1963.

\bibitem[Walter and Couzin(2021)]{walter2021trex}
Tristan Walter and Iain~D Couzin.
\newblock Trex, a fast multi-animal tracking system with markerless identification, and 2d estimation of posture and visual fields.
\newblock \emph{Elife}, 10:\penalty0 e64000, 2021.

\bibitem[Wojke et~al.(2017)Wojke, Bewley, and Paulus]{wojke2017simple}
Nicolai Wojke, Alex Bewley, and Dietrich Paulus.
\newblock Simple online and realtime tracking with a deep association metric.
\newblock In \emph{2017 IEEE international conference on image processing (ICIP)}, pages 3645--3649. IEEE, 2017.

\bibitem[Zhang et~al.(2022)Zhang, Sun, Jiang, Yu, Weng, Yuan, Luo, Liu, and Wang]{zhang2022bytetrack}
Yifu Zhang, Peize Sun, Yi Jiang, Dongdong Yu, Fucheng Weng, Zehuan Yuan, Ping Luo, Wenyu Liu, and Xinggang Wang.
\newblock Bytetrack: Multi-object tracking by associating every detection box.
\newblock In \emph{European Conference on Computer Vision}, pages 1--21. Springer, 2022.

\end{thebibliography}
}


\end{document}